\documentclass[runningheads]{llncs}
\usepackage[T1]{fontenc}
\usepackage{graphicx}
\usepackage{booktabs}
\usepackage[misc]{ifsym}
\newcommand{\corr}{(\Letter)}
\usepackage{amsmath, amssymb}
\usepackage{caption}
\usepackage[justification=centering]{subcaption}
\usepackage{pifont}

\usepackage{multirow}
\usepackage{xr-hyper}

\usepackage{makeidx}  % allows for indexgeneration
\usepackage{hyperref}
\usepackage{url}
\usepackage{graphicx}
\usepackage{algorithm}
\usepackage{algorithmic}
\usepackage{mathtools}
\usepackage{multirow}
\usepackage{caption}
\usepackage[justification=centering]{subcaption}
\usepackage{pdfpages}
\usepackage{pdflscape}
\usepackage{mathtools}
\usepackage{bbm}
\usepackage[T1]{fontenc}
\usepackage{lipsum}
\usepackage{amsmath, amssymb}
\usepackage{booktabs}
\newcommand*\rot[1]{\rotatebox{90}{#1}}

\usepackage{mwe}
\begin{document}

\title{Beyond Feature Fusion: Contextual Bayesian PEFT for Multimodal Uncertainty Estimation}

% Balanced Hierarchical Fusion for Long-Sequence Multimodal Representation Learning with Structure-Preserving and Information-Balanced Regularization
% Balanced Hierarchical Fusion for Long-Sequence Multimodal Representation Learning
% Balanced Hierarchical Fusion for Long-Sequence Audio-Text Representation Learning
% Beyond Pairwise Alignment: Reciprocal Contrastive Learning for Long-Sequence Audio-Text Representation
% Joint-Centric Dual Contrastive Learning with Structure Preservation and Information Balancing

\titlerunning{CoCo-LoRA: Contextual Bayesian PEFT for Uncertainty Estimation} %and Information-Balanced Regularization
% If the full title of your paper is short enough to also fit in the running head, you can omit the abbreviated paper title here. You can check as follows: if you comment out the \titlerunning line, something will appear in the header of all odd-numbered pages of your PDF from page 3 onward. This something is either the full title (in which case all is well), or the error message "Title Suppressed Due to Excessive Length". If this error message appears, you're going to want to provide an abbreviated title within the \titlerunning command, because if you won't do it, Springer will do it for you.

%N.B.: Author information (both in the \author{} and \authorrunning{} command) should only be present in the Camera-Ready Version of your paper. The version that you initially submit for review, ought to be double-blind. So, when initially submitting your paper, use:
% \author{Author information scrubbed for double-blind reviewing}

\author{Habibeh Naderi\inst{1} \corr \and
Behrouz Haji Soleimani\inst{1} \and
Stan Matwin\inst{1} %\orcidID{0000-1111-2222-3333}
}

% You may leave out the orcidID information, if you want to.
% Use \corr to indicate the corresponding author. Note the spacing around the \corr command. Only one author can be the corresponding author.

%N.B.: comment out the \authorrunning{} command for the double-blind version of your paper submitted for review. Later, if your paper is accepted, use the command for the Camera-Ready Version.

\authorrunning{H. Naderi et al.}

% If there are two authors, write 'A.L. Benjamin and C.C. Broadus Jr.'
% If there are more than two authors, '[...] et al.' is used.

\institute{Dalhousie University, Halifax NS, Canada \\ \email{habibeh.naderi@dal.ca, behrouz.hajisoleimani@dal.ca, stan@cs.dal.ca}
}

\maketitle              % typeset the header of the contribution

\begin{abstract}
We introduce CoCo-LoRA, a multimodal, uncertainty-aware parameter-efficient fine-tuning method for text prediction tasks accompanied by audio context. Existing PEFT approaches such as LoRA are efficient but typically deterministic, while recent Bayesian low-rank adapters model uncertainty in a lightweight way yet remain largely unimodal and condition uncertainty primarily on internal text features. This leaves them poorly equipped to reflect uncertainty driven by external acoustic factors such as background noise, channel variability, or speaking style, which can materially affect reliability in speech-centered applications. CoCo-LoRA addresses this gap by conditioning a contextual variational posterior in the low-rank space on both local text-derived adapter features and an audio-derived context signal. A pooled audio embedding is projected once into a shared context space and then adapted through lightweight layer-wise heads, enabling global-to-local, depth-specific modulation of the adapter uncertainty and update without high-dimensional multimodal fusion. Stochasticity is confined to a compact latent component in the rank space, preserving PEFT scalability while producing audio-sensitive, heteroscedastic uncertainty. Based on our evaluations across diverse tasks and backbone combinations, CoCo-LoRA consistently matches or outperforms text-only PEFT and conventional feature-fusion transfer baselines, particularly on high-coverage labels where reliable adaptation is critical. The results indicate that using audio as a contextual uncertainty signal, rather than as a fused feature stream, provides a robust and parameter-efficient alternative for multimodal low-resource prediction.

\end{abstract}

\section{Introduction}
\label{sec:intro_gpt}

Pre-trained Large Language Models (LLMs) have become the dominant backbone for a wide range of prediction tasks \cite{llms_brown2020language,llms_radford2019language,llms_minaee2024large,llms_singhal2023publisher}, but adapting them to specialized domains remains challenging when labeled data are limited and the application is high-stakes \cite{naderi2020generating}. This tension is especially pronounced in clinical and behavioral settings, where downstream targets are heterogeneous (e.g., affect, cognition, interaction quality), annotations are expensive, and model confidence is as important as raw accuracy. In such regimes, practitioners must balance three competing requirements: (i) strong task performance under limited supervision, (ii) computational and memory efficiency during adaptation, and (iii) reliable uncertainty estimates that reflect when the model should be trusted.

Parameter-efficient fine-tuning (PEFT) \cite{fu2023effectiveness_peft} addresses the first two requirements by updating only a small set of adapter parameters while keeping the backbone frozen. Methods such as LoRA and its variants (e.g., low-rank adapters, structured low-rank factorizations, and other lightweight modules) enable fast adaptation with a fraction of trainable parameters compared to full fine-tuning, often retaining competitive performance \cite{hu2022lora,hyeon2021fedpara_loha,zhang2023adaptive_adalora,liu2022few_ia3}. However, standard PEFT mechanisms are typically deterministic and optimized primarily for accuracy, which can lead to overconfident predictions in distribution shifts, noisy inputs, or ambiguous examples \cite{uncertainty_leng2024taming,uncertainty_xiong2023can}. This limitation motivates uncertainty-aware PEFT approaches \cite{wang2024blob,rahmati2025clora,laplacelora_yang2023bayesian,samplelora_fan2021contextual} that explicitly model epistemic and/or heteroscedastic uncertainty, especially in settings where errors are costly.

Heteroscedastic aleatoric uncertainty captures input-dependent noise inherent in the data distribution, whereas epistemic uncertainty reflects model uncertainty arising from limited data and parameter uncertainty, and can in principle be reduced with additional training data.

\begin{figure}[t]
\centering
\includegraphics[width=0.99\textwidth]{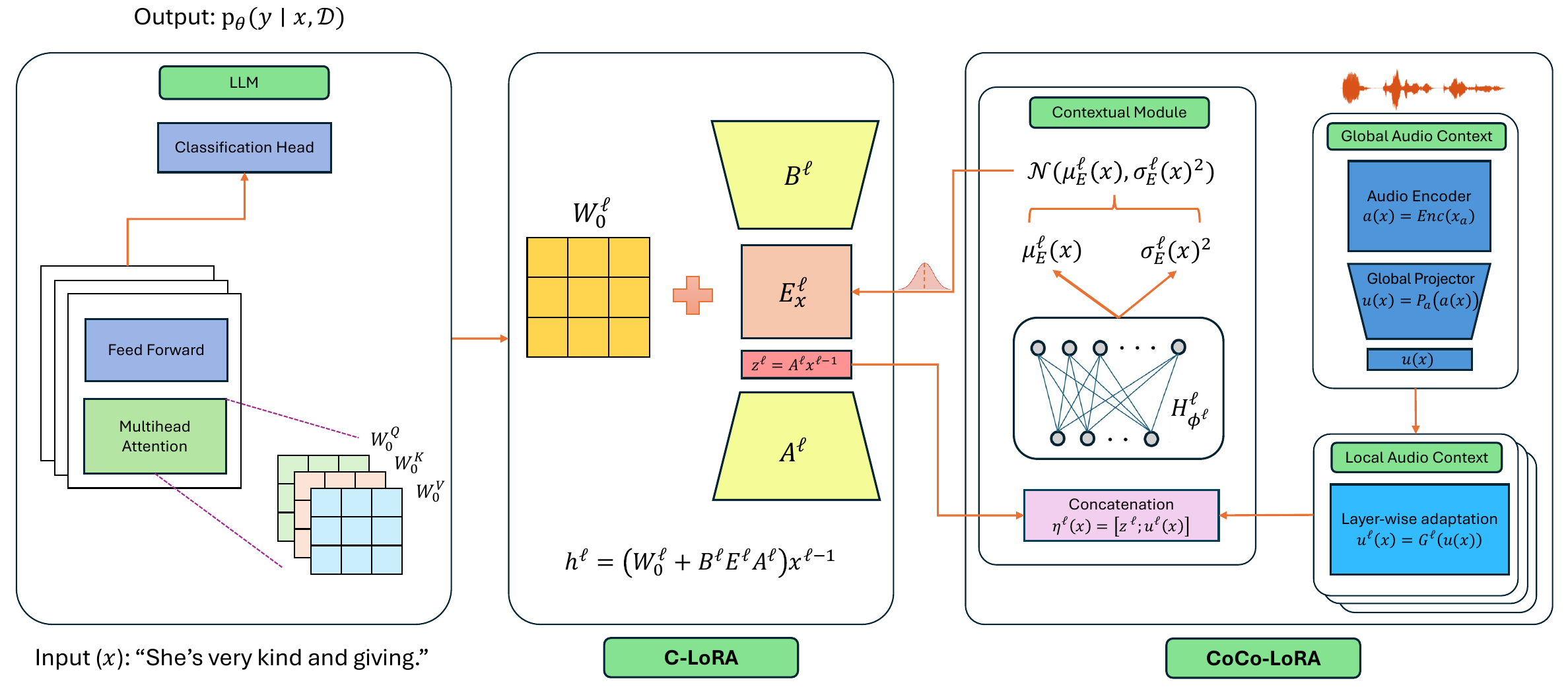}
\caption{\label{fig_architecture}Overview of the proposed CoCo-LoRA Architecture. Global and layer-wise audio context, condition the adapters through a variational Gaussian module, enabling uncertainty-aware multimodal low-rank adaptation.}
\end{figure}

A second, orthogonal challenge arises when the predictive signal is inherently multimodal. For many human-centered prediction tasks, speech acoustics and lexical content provide complementary evidence. Transfer learning \cite{domainadaptation_ben2006analysis} studies repeatedly show that text often dominates aggregate performance, while audio can provide targeted gains when prosody, speaker state, or environmental conditions contain discriminative cues \cite{cpc_oord2018representation,audiopalm_rubenstein2023audiopalm,audiolm_borsos2023audiolm}. The co-occurrence of multimodal observations acts as a form of weak supervision \cite{weaklysupwo_locatello2020weakly}, where paired data points serve as weak labels indicating an underlying shared but unobserved ground truth. Through the cooperation between the two modalities, multimodal learning can achieve better feature learning, leading to improvements in performance on downstream tasks compared to unimodal learning \cite{naderi2019multimodal,Naderi2025MAC,pmlr-naderi2026coprime}. Yet multimodal modeling introduces additional complexity: feature fusion can increase effective dimensionality, amplify over-parameterization in low-data regimes, and complicate calibration when modalities disagree. These issues are exacerbated when the audio is noisy or recording conditions vary widely, exactly the cases where meaningful uncertainty is most valuable.

Recent research has explored uncertainty-aware adapters in the LoRA family by Bayesianizing parts of the adaptation mechanism. Bayesian LoRA by Backpropagation (BLoB) \cite{wang2024blob} introduces a Bayesian treatment for LoRA-style parameters using Variational Inference (VI), capturing epistemic uncertainty in adapter weights via an input-independent posterior. Contextual Low-Rank Adaptation (C-LoRA) \cite{rahmati2025clora} advances this direction by making uncertainty data-dependent: it proposes a contextual variational posterior in the low-rank space, enabling sample-specific (heteroscedastic) uncertainty while keeping the Bayesian complexity lightweight. Despite this progress, existing approaches are unimodal and +largely condition uncertainty on internal language-model features alone. As a result, they do not explicitly account for uncertainty induced by external modalities (e.g., background noise, speaking style, prosodic variability), even though such factors can materially affect the reliability of predictions in speech-centered applications.

In this work, we introduce \textbf{Co}ntext-\textbf{Co}nditioned \textbf{Lo}w-\textbf{R}ank \textbf{A}daptation (\textbf{CoCo-LoRA}), a multimodal, uncertainty-aware PEFT framework designed for text prediction tasks accompanied by audio context (Figure \ref{fig_architecture}). CoCo-LoRA builds on the key scalability insight of C-LoRA: rather than Bayesianizing large weight matrices or full LoRA factors, we confine stochasticity to a lightweight latent component in the rank-$r$ space, preserving the computational efficiency that makes PEFT practical. The central novelty is that we condition the variational posterior not only on local low-rank text features but also on a shared global audio context derived from a pooled audio embedding. This context is injected into the uncertainty pathway through a layer-wise adaptation mechanism, allowing different transformer layers to attend to audio-driven cues in a depth-specific way without incurring the heavy parameter overhead of full multimodal fusion.

Conceptually, CoCo-LoRA bridges three lines of work: PEFT for efficient adaptation, uncertainty-aware LoRA for principled confidence modeling, and multimodal transfer learning for leveraging complementary acoustic information. From the PEFT perspective, CoCo-LoRA retains the practical benefits of frozen backbones and small trainable modules, making it suitable for limited-resource training and rapid experimentation across many tasks. From the uncertainty modeling perspective, it inherits amortized, input-dependent uncertainty from contextual Bayesian adapters, while extending it to be explicitly sensitive to cross-modal information. From the multimodal perspective, CoCo-LoRA offers a lightweight alternative to early/late fusion strategies: rather than expanding the classifier or backbone with high-dimensional fused representations, it uses audio to modulate the distribution over adapter behavior, enabling the model to become more cautious when audio context indicates unreliable conditions and more confident when the acoustic signal is consistent with the text.

CoCo-LoRA is also complementary to representation learning approaches commonly used in multimodal settings. Contrastive objectives \cite{clip_radford2021learning,simclr_chen2020simple,simclr2_chen2020big} (e.g., aligning audio and text embeddings) and transfer learning strategies (e.g., frozen encoders with task-specific heads) have demonstrated strong performance in low-label regimes by encouraging modality-invariant or modality-aligned representations. CoCo-LoRA does not replace these paradigms, instead, it provides an orthogonal mechanism: it uses audio not primarily to increase separability through feature fusion, but to shape uncertainty and adaptation in a principled Bayesian manner. This design is particularly appealing for human-centered prediction tasks where the model must handle dataset shifts across speakers, recording environments, and interaction dynamics. This makes CoCo-LoRA a very well aligned method for clinical interview analysis.

Overall, CoCo-LoRA aims to deliver a practical answer to a recurring problem in multimodal, low-resource prediction: how to adapt large language models efficiently while producing confidence estimates that respond to audio-driven reliability cues, without paying the parameter and optimization costs of full multimodal fine-tuning. Our contributions can be outlined as follows:

% \paragraph{Contributions.}
\begin{itemize}
    \item We propose CoCo-LoRA, a multimodal, uncertainty-aware PEFT framework that conditions adapter uncertainty on external audio context in addition to internal low-rank text features, enabling audio-sensitive heteroscedastic uncertainty.
    \item We introduce a global-to-local context pathway for audio: a single pooled audio embedding is projected into a shared context space and then adapted per layer with lightweight heads, allowing depth-specific specialization without high-dimensional multimodal fusion.
    \item We preserve lightweight Bayesian complexity by restricting stochasticity to a compact latent component in the rank-$r$ space, maintaining the scalability advantages of contextual LoRA-style Bayesianization.
    \item We provide a unified perspective connecting PEFT, uncertainty-aware LoRA (e.g., BLoB and C-LoRA), and multimodal transfer learning, highlighting how cross-modal context can be incorporated as a principled uncertainty signal rather than only a feature-fusion signal.
\end{itemize}

\section{Proposed Method}
\label{sec:coco_lora}

\subsection{Preliminaries}

\paragraph{\textbf{Low-Rank Adaptation (LoRA)}}
Low-Rank Adaptation (LoRA) is a parameter-efficient fine-tuning method designed to adapt large pre-trained models to downstream tasks while keeping the backbone weights frozen \cite{hu2022lora}. The core assumption behind LoRA is that task-specific weight updates lie in a low-dimensional subspace. Instead of updating the full weight matrix, LoRA learns a low-rank decomposition of the weight update.

Consider a linear transformation with frozen pre-trained weights $W_0 \in \mathbb{R}^{d \times k}$. The adapted forward pass is defined as:
\begin{equation}
h \;=\; (W_0 + \Delta W)x \;=\; (W_0 + BA)x,
\label{eq:lora_standard}
\end{equation}
where $x \in \mathbb{R}^{k}$ is the input, $h \in \mathbb{R}^{d}$ is the output, and $\Delta W = BA$ is the low-rank update. Here, $B \in \mathbb{R}^{d \times r}$ and $A \in \mathbb{R}^{r \times k}$ are trainable matrices with rank $r \ll \min(d,k)$. 

This factorization reduces the number of trainable parameters from $d\times k$ to $r\times(d+k)$, yielding substantial savings in memory and computation while often matching full fine-tuning performance.

\paragraph{\textbf{Bayesian Uncertainty-Aware Low-Rank Adaptation}}
While LoRA provides efficient deterministic adaptation, it does not model uncertainty in the learned parameters. Full Bayesian inference over all backbone weights is computationally infeasible for modern transformer models. A scalable alternative is to restrict uncertainty modeling to the low-rank adapters.

Bayesian LoRA by Backpropagation (BLoB) introduces uncertainty into the LoRA parameters while keeping the frozen backbone intact. Given the LoRA decomposition $\Delta W = BA$, BLoB maintains $B$ as deterministic and places a variational posterior over $A$. Specifically, a mean-field variational inference approximation is used: $q(A) = \mathcal{N}\big(A \mid \mu_A, \Omega_A^2\big)$, where $\mu_A$ and $\Omega_A^2$ denote the variational mean and variance parameters. The variational parameters are learned by maximizing the Evidence Lower BOund (ELBO):
\[
\mathcal{L}' 
= \mathbb{E}_{q} \big[ \log p(\mathcal{D} \mid A, B) \big]
- \mathrm{KL}\big[q(A) \,\|\, p(A)\big].
\]

The first term corresponds to the expected log-likelihood under the variational posterior, while the second term regularizes the solution via the Kullback-Leibler divergence to the prior. Optimization is performed using the reparameterization trick, enabling gradient-based training of both mean and variance parameters.

By confining Bayesian inference to the low-rank adapters, BLoB preserves the scalability benefits of LoRA while providing principled uncertainty estimation. This makes it an effective compromise between deterministic fine-tuning and fully Bayesian neural networks.

\subsection{Context-Conditioned Low-Rank Adaptation (CoCo-LoRA)}

The preliminaries above highlight a practical gap in low-resource, high-stakes settings. Deterministic PEFT (e.g., LoRA) is efficient, but it can be overconfident when supervision is scarce or inputs are ambiguous. Bayesianizing adapters (e.g., BLoB) introduces epistemic uncertainty, yet its posterior is typically input-independent, which limits its ability to react to sample-specific noise. C-LoRA addresses this by amortizing a contextual variational posterior in the low-rank space, producing heteroscedastic (input-dependent) uncertainty, but it conditions that uncertainty primarily on internal language-model features. In multimodal speech-centered applications, however, uncertainty is often driven by external factors (e.g., background noise, channel variability, speaker state) that are visible in audio but not necessarily recoverable from text alone. 

CoCo-LoRA extends contextual Bayesian adapters to the multimodal regime by letting audio context shape the variational posterior over the lightweight latent in the rank-$r$ space. A visual representation of our CoCo-LoRA framework is presented in Figure \ref{fig_architecture}. The key design principle is: use audio as a reliability signal, not as heavy feature fusion. Concretely, we keep the transformer backbone frozen and avoid inflating representation dimensionality. Instead, we route a compact global audio embedding through a global-to-local pathway that modulates per-layer uncertainty and adaptation.

\paragraph{\textbf{Setup and notation.}}
Let $\mathcal{D}=\{(x_i,y_i)\}_{i=1}^N$ denote a dataset of $N$ i.i.d.\ samples. We consider a pre-trained transformer with $L$ layers. In a linear sub-layer at layer $\ell\in[L]$, let $W^\ell_0\in\mathbb{R}^{d\times k}$ be the frozen pre-trained weight, $x^{\ell-1}\in\mathbb{R}^{d}$ the input activation, and $h^\ell\in\mathbb{R}^{d}$ the output.
LoRA parameterizes a low-rank update $\Delta W^\ell$ via matrices $B^\ell\in\mathbb{R}^{d\times r}$ and $A^\ell\in\mathbb{R}^{r\times k}$ with rank $r\ll d$. Following the lightweight LoRA factorization introduced in C-LoRA \cite{rahmati2025clora}, we insert an additional matrix $E^\ell$ in the low-dimensional space:
\begin{equation}
h^\ell \;=\; \bigl(W^\ell_0 + \Delta W^\ell(x)\bigr)x^{\ell-1}
\;=\; W^\ell_0 x^{\ell-1} + B^\ell E^\ell_x A^\ell x^{\ell-1},
\qquad
\Delta W^\ell(x)\;=\;B^\ell E^\ell_x A^\ell .
\label{eq:coco_forward}
\end{equation}
LoRA intermediate feature is also defined as: $z^\ell = A^\ell x^{\ell-1}\in\mathbb{R}^{r}$.

\paragraph{\textbf{Where uncertainty lives and what audio controls.}}
CoCo-LoRA restricts stochasticity to the compact latent $E^\ell_x\in\mathbb{R}^{r\times r}$, as in C-LoRA, keeping $A^\ell$ and $B^\ell$ deterministic. This preserves the scalability of PEFT while enabling uncertainty estimates through variational inference. The distinction from prior work is what the posterior conditions on: instead of deriving context solely from internal representations, we incorporate an external audio context so that uncertainty can increase when acoustics suggest unreliability (e.g., noisy segments) and decrease when the signal is clean and consistent.

\paragraph{\textbf{Global audio context.}}
Each input sample $x$ is accompanied by a pooled audio embedding $a(x)\in\mathbb{R}^{d_a}$ obtained from a (frozen or separately trained) audio encoder. CoCo-LoRA maps $a(x)$ into a shared context space of dimension $c\ll d_a$ via a learned projection $P_a:\mathbb{R}^{d_a}\to\mathbb{R}^{c}$:
\begin{equation}
u(x) \;=\; P_a(a(x))\in\mathbb{R}^{c}.
\label{eq:coco_audio_proj}
\end{equation}
This yields a single global context vector $u(x)$ that is inexpensive to compute and share across all layers, providing a consistent audio-derived reliability cue that can influence uncertainty throughout the transformer.

\paragraph{\textbf{Layer-wise adaptation of audio context.}}
A global context may affect different depths in different ways: lower layers may benefit from coarse acoustic cues, while higher layers may require task-specific, abstract signals. CoCo-LoRA therefore uses small per-layer heads to convert the shared context into a layer-specific vector in the rank-$r$ space:
\begin{equation}
\tilde u^\ell(x) \;=\; G^\ell\!\bigl(u(x)\bigr)\in\mathbb{R}^{r}.
\label{eq:coco_layer_audio}
\end{equation}
This global-to-local pathway preserves efficiency (shared projection once per sample) while allowing each layer to specialize its use of audio without resorting to high-dimensional multimodal fusion.

\paragraph{\textbf{Context-conditioned variational posterior over $\mathbf{E^\ell_x}$.}}
We model $E^\ell_x\in\mathbb{R}^{r\times r}$ as a latent random matrix with an input-dependent Gaussian variational posterior. The posterior is conditioned on (i) the local LoRA feature $z^\ell$ capturing how the current text activation enters the low-rank adapter, and (ii) the audio-conditioned layer context $\tilde u^\ell(x)$ capturing external reliability information. We concatenate these to form the contextual summary:
\begin{equation}
\eta^\ell(x) \;=\; [\,z^\ell;\tilde u^\ell(x)\,]\in\mathbb{R}^{2r}.
\label{eq:coco_concat}
\end{equation}
Let $H^\ell_{\phi^\ell}:\mathbb{R}^{2r}\to\mathbb{R}^{2r^2}$ output a mean vector $\mu^\ell_E(x)\in\mathbb{R}^{r^2}$ and a log-variance vector $\log v^\ell_E(x)\in\mathbb{R}^{r^2}$. Then
\begin{equation}
q_{\phi^\ell}\!\left(\mathrm{vec}(E^\ell_x)\mid x\right)
\;=\;
\mathcal{N}\!\left(\mu^\ell_E(x),\,\mathrm{diag}\!\bigl(\sigma^\ell_E(x)^2\bigr)\right),
\label{eq:coco_q}
\end{equation}
where $\sigma^\ell_E(x)\in\mathbb{R}^{r^2}_{>0}$ is parameterized using a scaled softplus:
\begin{equation}
\sigma^\ell_E(x)
\;=\;
\varepsilon\,\mathrm{softplus}\!\bigl(\log v^\ell_E(x)\bigr) + \delta,
\qquad
\varepsilon>0,\;\delta\approx 10^{-6}.
\label{eq:coco_sigma}
\end{equation}
Eq.~\eqref{eq:coco_q} induces heteroscedastic uncertainty because both $\mu^\ell_E(x)$ and $\sigma^\ell_E(x)$ depend on $x$ through $z^\ell$ and the audio-conditioned context $\tilde u^\ell(x)$. This is precisely what allows CoCo-LoRA to react to cross-modal reliability cues: the same text pattern can yield different posterior variance depending on the accompanying audio context.

\paragraph{\textbf{Reparameterized sampling and stochastic low-rank update.}}
We sample $E^\ell_x$ via the reparameterization trick. Let $\xi^\ell\sim\mathcal{N}(0,I_{r^2})$:
\begin{equation}
\mathrm{vec}(E^\ell_x)
\;=\;
\mu^\ell_E(x) + \sigma^\ell_E(x)\odot \xi^\ell,
\label{eq:coco_reparam}
\end{equation}
and reshape to $E^\ell_x\in\mathbb{R}^{r\times r}$. The resulting stochastic adapter contribution is
\begin{equation}
\Delta h^\ell(x)
\;=\;
B^\ell E^\ell_x z^\ell
\;=\;
B^\ell E^\ell_x A^\ell x^{\ell-1}.
\label{eq:coco_delta_h}
\end{equation}
Importantly, stochasticity is confined to $r\times r$ variables, keeping Bayesian complexity per layer at $O(r^2)$ rather than scaling with the backbone dimension $d$.

\paragraph{\textbf{Prior and ELBO objective.}}
We place a factorized Gaussian prior on $\mathrm{vec}(E^\ell)$:
\begin{equation}
p\!\left(\mathrm{vec}(E^\ell)\right)
\;=\;
\mathcal{N}\!\left(0,\beta^2 I_{r^2}\right),
\label{eq:coco_prior}
\end{equation}
where $\beta>0$ controls the prior scale. Let $\theta$ collect all deterministic parameters (including $\{A^\ell,B^\ell\}_{\ell=1}^L$ and the audio projections/adapters), and let $\phi=\{\phi^\ell\}_{\ell=1}^L$ denote variational parameters of the inference heads.
We maximize the ELBO:
\begin{equation}
\mathcal{L}(\theta,\phi)
=
\sum_{i=1}^N
\Biggl[
\mathbb{E}_{q_\phi(E_{x_i}\mid x_i)}
\log p_\theta\!\left(y_i\mid x_i, E_{x_i}\right)
-
\gamma\sum_{\ell=1}^L
\mathrm{KL}\!\left(
q_{\phi^\ell}\!\left(E^\ell_{x_i}\mid x_i\right)
\,\|\, p(E^\ell)
\right)
\Biggr],
\label{eq:coco_elbo}
\end{equation}
with $\gamma>0$ a KL reweighting coefficient. The complete learning objective is expressed as a summation over all samples: $\mathcal{L}= \sum_{(x,y)\in\mathcal{D}}\mathcal{L}(x,y)$. Then, we learn the deterministic parameters B and A together represented by $\theta$ by minimizing the expected negative log-likelihood (NLL). The KL term keeps the contextual posterior close to the prior unless the data support deviating from it. The conditioning on $\tilde u^\ell(x)$ means that this deviation can systematically depend on audio-indicated reliability.

\paragraph{\textbf{Stochastic optimization.}}
Using one Monte Carlo draw per sample (as in C-LoRA), we approximate the expected log-likelihood term as:
\begin{equation}
\mathbb{E}_{q_\phi(E_{x_i}\mid x_i)}
\log p_\theta\!\left(y_i\mid x_i,E_{x_i}\right)
\;\approx\;
\log p_\theta\!\left(y_i\mid x_i,E^{(1)}_{x_i}\right),
\qquad
E^{(1)}_{x_i}\sim q_\phi(E_{x_i}\mid x_i).
\label{eq:coco_mc_train}
\end{equation}

\paragraph{\textbf{Posterior predictive inference and audio-sensitive uncertainty.}}
At test time, CoCo-LoRA supports either a deterministic approximation using posterior means or Monte Carlo model averaging. For a test input $x^\ast$ with audio embedding $a(x^\ast)$:
\begin{equation}
p(y^\ast\mid x^\ast,\mathcal{D})
\;\approx\;
\frac{1}{M}\sum_{m=1}^M
p_\theta\!\left(y^\ast\mid x^\ast, E^{(m)}_{x^\ast}\right),
\qquad
E^{(m)}_{x^\ast}\sim q_\phi(E_{x^\ast}\mid x^\ast).
\label{eq:coco_ppd}
\end{equation}
Because $q_\phi(E_{x^\ast}\mid x^\ast)$ is conditioned on both $z^\ell$ and the audio context $u(x^\ast)$ (via $\tilde u^\ell(x^\ast)$), predictive uncertainty becomes explicitly audio-sensitive and heteroscedastic. Practically, this yields a principled mechanism for modulating confidence based on acoustic context without changing the backbone or introducing heavy multimodal fusion blocks.

\paragraph{\textbf{Summary of novelties and differences to prior work.}}
CoCo-LoRA preserves C-LoRA's lightweight Bayesianization by restricting stochasticity to $E^\ell_x\in\mathbb{R}^{r\times r}$, keeping $A^\ell$ and $B^\ell$ deterministic and avoiding the $O(d)$-scaled Bayesian overhead typical of Bayesianizing full LoRA factors.
Unlike BLoB, the posterior is amortized and input-dependent, yielding sample-specific (heteroscedastic) uncertainty.
Unlike C-LoRA, the contextual posterior is conditioned not only on internal low-rank features $z^\ell$ but also on a shared external audio context, enabling uncertainty to reflect cross-modal, audio-driven characteristics while maintaining parameter efficiency.

\section{Dataset}
\label{sec_data}

\begin{table}[tp]
\centering
\setlength\tabcolsep{2pt}
\scalebox{0.84}{
\begin{tabular}{llcccccccc}
\toprule
 & \textbf{Task} & \textbf{\# Samples} & \textbf{\# Classes} & \textbf{Imbalanced (\%)} & \textbf{Label 0} & \textbf{Label 1} & \textbf{Label 2} & \textbf{Label 3} \\
 % & \textbf{Task} & \rot{\textbf{Num Samples}} & \rot{\textbf{Num Classes}} & \rot{\textbf{Imbalanced (\%)}} & \rot{\textbf{Label 0}} & \rot{\textbf{Label 1}} & \rot{\textbf{Label 2}} & \rot{\textbf{Label 3}} \\
\midrule
Parent & objective & 23693 & 2 & 19.08 & 13096 & 10597 & - & - \\
Parent & sentiment & 23479 & 3 & 57.39 & 4725 & 11088 & 7666 & - \\
Parent & anger & 23694 & 2 & 92.22 & 21984 & 1710 & - & - \\
Parent & fear & 23694 & 2 & 95.29 & 22628 & 1066 & - & - \\
Parent & joy & 23690 & 2 & 66.58 & 17756 & 5934 & - & - \\
Parent & sadness & 23694 & 2 & 92.70 & 22081 & 1613 & - & - \\
Parent & neutral & 23693 & 2 & 16.28 & 10797 & 12896 & - & - \\
Parent & cohesion & 23694 & 2 & 81.32 & 19964 & 3730 & - & - \\
Parent & rumination & 23694 & 2 & 98.95 & 23447 & 247 & - & - \\
Parent & overinclusive & 23694 & 2 & 97.20 & 23049 & 645 & - & - \\
Parent & worry & 23694 & 2 & 91.76 & 21890 & 1804 & - & - \\
Parent & criticism & 23694 & 2 & 88.42 & 21235 & 2459 & - & - \\
\midrule
Offspring & objective & 10319 & 2 & 1.90 & 5209 & 5110 & - & - \\
Offspring & sentiment & 10222 & 3 & 52.51 & 2404 & 5062 & 2756 & - \\
Offspring & richness & 10273 & 3 & 97.08 & 1366 & 8654 & 253 & - \\
Offspring & reference & 10328 & 4 & 77.89 & 2072 & 4970 & 2187 & 1099 \\
Offspring & irrelevance & 10301 & 2 & 87.98 & 9196 & 1105 & - & - \\
Offspring & anger & 10329 & 2 & 95.55 & 9889 & 440 & - & - \\
Offspring & fear & 10329 & 2 & 99.04 & 10231 & 98 & - & - \\
Offspring & joy & 10329 & 2 & 94.49 & 9790 & 539 & - & - \\
Offspring & sadness & 10329 & 2 & 94.62 & 9802 & 527 & - & - \\
Offspring & neutral & 10328 & 2 & 80.92 & 1655 & 8673 & - & - \\
Offspring & coherence & 10329 & 2 & 68.35 & 7846 & 2483 & - & - \\
Offspring & rumination & 10329 & 2 & 99.01 & 10228 & 101 & - & - \\
Offspring & worry & 10329 & 2 & 99.13 & 10240 & 89 & - & - \\
Offspring & anxiousness & 10317 & 2 & 96.51 & 9969 & 348 & - & - \\
Offspring & aggression & 2182 & 2 & 99.49 & 2171 & 11 & - & - \\
Offspring & criticism & 8090 & 2 & 95.04 & 7708 & 382 & - & - \\
Offspring & self-criticism & 10329 & 2 & 96.77 & 10006 & 323 & - & - \\
\bottomrule
\end{tabular}
}
\caption{Dataset information for the parent and offspring data.}
\label{tbl_dataset_info}
\end{table}

The data used in this work consists of audio speech samples from 369 subjects participating in the Families Overcoming Risks and Building Opportunities for Well Being (FORBOW) research project \cite{forbow_uher2014familial,naderi2019multimodal}. Participants are parents, 266 mothers and 103 fathers, in the age range of 28-51 years. In these clinical interviews, parents were asked to talk about their children for five minutes without interruption. Out of these subjects, 149 were diagnosed with Major Depressive Disorder (MDD), 66 were diagnosed with Bipolarity Disorder (BD), 19 were diagnosed with Schizophrenia, and 129 were the control group with no major mood disorders.

In addition to the parents' interview files, FORBOW research project collected interviews with the children themselves. The audio interviews of children consists of 3 parts: 1) a three minute interview where children talk about themselves, 2) a two minute interview talking about a positive experience they had, and 3) a two minute interview where they talk about a negative experience they had. All three interviews are uninterrupted with a total of 7 minutes of speech from each child. 

% We transcribed and broke down each sample into multiple segments based on changes in emotion, sentiment, objectivity/subjectivity, etc. Average word count in a segment is 17 and average audio length for a segment is 6.47 seconds. A segment has been labeled as subjective if it includes expression of opinion, beliefs, or personal thoughts of the speaker. In contrary, if the segment consists of facts or observations of the speaker, it has been labeled as objective. Four basic emotions are considered in this analysis including anger, fear, joy, and sadness. Children segments also have some additional labels such as self-criticism, irrelevance, aggression, anxiousness, richness, etc. There are a total of 17 segment-level tasks for the children files. Six multidisciplinary researchers rated each segment for sentiment, subjectivity, emotions (anger, fear, joy, sadness, neutral), cohesion, rumination, over-inclusiveness, worry, and criticism. 5,818 segments were rated by two or more researchers and the intraclass correlation for ratings of different researchers was high showing strong agreement in the labeling. Table \ref{tbl_dataset_info} show detailed information about the dataset.

We transcribed and broke down each sample into multiple segments based on changes in emotion, sentiment, objectivity/subjectivity, etc \cite{naderi2018deepminer}. Average word count in a segment is 17 and average audio length for a segment is 6.47 seconds. Six multidisciplinary researchers rated these segments for subjectivity, emotions, sentiment, criticism, etc. 5,818 segments were rated by two or more researchers and the intraclass correlation for ratings of different researchers was high showing strong agreement in the labeling. Table \ref{tbl_dataset_info} show detailed information about the dataset. For each task, the imbalanced \% is calculated as $100\times(\max_{c\in C}n_c - \min_{c\in C}n_c) / \max_{c\in C}n_c$. As we can see most of the tasks are quite imbalanced.

% Figure \ref{fig_mood} illustrates the mood, emotions, and sentiment changes in a random sample from parent data. We can also see sample segments shown at the top of the figure which gives an idea of how the segmentation and the labeling is done. All text excerpts are paraphrased for privacy.

% \begin{figure}[tp]
% \centering
% \includegraphics[width=0.9\textwidth]{./figs/dataset/mood_parent_800-0036-060_M1_2017_CH_ch.pdf}
% \caption{\label{fig_mood}Emotion and sentiment changes in a random sample.}
% \end{figure}

\section{Experimental Results}
\label{sec:results}

In this section we provide the experimental results of our proposed context conditioned fine-tuning method on the aforementioned dataset. For the experiments in this section, we chose 4 different language models, nliRoBERTa, paraTinyBERT, allMiniLM12, and allDistilRoBERTa. We also chose three pre-trained audio models, whisperMedium, wav2vec2Large-FineTune, and hubertLargeFineTune. This results in a total of 12 combinations of backbones for CoCo-LoRA fine-tuning that are trained separately on each individual task for both parent and offspring data. All results in this section are based on a 5-fold cross-validation.

We compare three families of approaches: 
\begin{enumerate}
    \item Text-only PEFT such as LoRA and C-LoRA that adapt the frozen LLM using low-rank updates on text only (no audio).
    \item Multimodal transfer learning with feature fusion: we obtain a pooled audio embedding from an audio encoder and fuse it with the text representation (e.g., concatenation), followed by a task-specific classification head (3-layer MLP, 32 units followed by 16 units followed by a softmax). This is a conventional lightweight multimodal baseline that injects audio at the representation/head level.
    \item CoCo-LoRA (ours): instead of heavy feature fusion, we use audio as context that conditions the variational low-rank adapter uncertainty and adaptation throughout the transformer. This preserves parameter-efficiency while making predictions (and uncertainty) explicitly audio-sensitive.
\end{enumerate}

For all PEFT techniques including LoRA, C-LoRA, and our CoCo-LoRA we used rank of 8, $r=8$, $\alpha=32$, and two epochs of training throughout the experiments. For CoCo-Lora, the shared audio down projector is an MLP that maps audio embeddings into 16 dimensions. And the layer-wise adaptation heads are single linear layers mapping the global context into 8 dimensions matching the $z^\ell$ dimensionality. Other parameters are set as follows: $\beta=0.2, \varepsilon=0.05, \gamma=0.008$. AdamW with learning rate of 0.001 and weight decay of 0.001 is used as optimizer. For inference, we use the average of 10 Monte Carlo samples, $M=10$.

Table \ref{tbl_parent} shows a consistent pattern across language-model backbones and audio encoders: CoCo-LoRA is usually the best method or among the top performers, with the most consistent gains on the two high-coverage labels, objective and sentiment. This suggests that when supervision is sufficient for the model to learn robust patterns, using audio to condition the adapter’s uncertainty and adaptation provides a stronger and more reliable signal than either text-only PEFT or a standard multimodal transfer baseline that fuses audio at the representation/head level.

A notable pattern is that CoCo-LoRA often surpasses feature-fusion transfer even when both methods use the same text backbone and the same audio encoder. This indicates that simply concatenating an audio embedding into the classifier is not always the most effective way to exploit acoustics in a low-resource, imbalanced setting. Instead, feeding audio through the uncertainty/adaptation pathway lets the model treat acoustics as a context or reliability cue that shapes how the backbone is adapted, without expanding the representation or relying on a large multimodal head that can overfit.

Against text-only baselines, CoCo-LoRA generally matches or improves upon LoRA on most tasks, and it improves over C-LoRA in many cases. This supports the central hypothesis behind the method: part of the uncertainty relevant to speech-centered predictions comes from external factors visible in audio, and conditioning the contextual posterior on audio context captures information that may not be recoverable from text features alone. The effect is not dependent on one particular audio model, the overall trend holds across Whisper, wav2vec2, and HuBERT, although the best encoder can vary by task.

CoCo-LoRA is not uniformly dominant on the most extreme class-imbalance labels, such as rumination and overinclusive. In these cases, the feature-fusion transfer baseline sometimes achieves higher AUC, which may reflect its ability to over-specialize to a small set of strong acoustic correlates in rare-positive regimes. By contrast, CoCo-LoRA’s variational objective regularizes adapter deviations via the KL term, which can make updates more conservative when evidence is sparse. Even so, CoCo-LoRA remains competitive on these tasks while offering a stronger overall pattern of improvements across tasks, backbones, and audio encoders.

\begin{table}[!t]
\centering
\setlength\tabcolsep{2.75pt}
\scalebox{0.79}{
\hspace{-2.0cm}
\begin{tabular}{lllcccccccccccc}
\toprule
 % &  &  & \textbf{obj} & \textbf{sentim} & \textbf{anger} & \textbf{fear} & \textbf{joy} & \textbf{sad} & \textbf{neutral} & \textbf{cohesion} & \textbf{rumination} & \textbf{overinclusive} & \textbf{worry} & \textbf{criticism} \\
 & & & \rot{\textbf{objective}} & \rot{\textbf{sentiment}} & \rot{\textbf{anger}} & \rot{\textbf{fear}} & \rot{\textbf{joy}} & \rot{\textbf{sadness}} & \rot{\textbf{neutral}} & \rot{\textbf{cohesion}} & \rot{\textbf{rumination}} & \rot{\textbf{overinclude}} & \rot{\textbf{worry}} & \rot{\textbf{criticism}} \\
Method & LLM & Audio Encoder &  &  &  &  &  &  &  &  &  &  &  &  \\
\midrule
LoRA & nRoBERTa & - & 88.02 & 92.38 & 90.87 & 89.81 & 89.77 & 89.92 & 79.49 & 78.30 & 59.01 & 85.76 & 89.99 & 91.93 \\
C-LoRA & nRoBERTa & - & 86.82 & 91.42 & 88.76 & 75.32 & 87.74 & 87.95 & 77.39 & 66.40 & 55.68 & 65.06 & 86.94 & 88.91 \\
\cmidrule[0.05pt](lr){1-3}
\multirow[t]{3}{*}{Transfer} & \multirow[t]{3}{*}{nRoBERTa} & whisperM & 85.70 & 90.39 & 89.56 & 89.10 & 89.13 & 89.51 & 78.28 & 79.63 & 82.40 & 83.55 & 89.03 & 90.15 \\
 &  & wav2vec2LFT & 85.15 & 91.07 & 89.20 & 88.43 & 88.69 & 88.76 & 77.36 & 76.89 & 79.16 & 82.25 & 88.47 & 90.15 \\
 &  & hubertLFT & 85.08 & 90.94 & 88.31 & 87.87 & 88.56 & 88.95 & 76.92 & 77.06 & 79.70 & 82.31 & 88.26 & 89.67 \\
\cmidrule[0.05pt](lr){1-3}
\multirow[t]{3}{*}{CoCo-LoRA} & \multirow[t]{3}{*}{nRoBERTa} & whisperM & 90.03 & 94.86 & 89.84 & 88.52 & 91.65 & \textbf{91.52} & 80.53 & 78.15 & 59.43 & 82.45 & \textbf{91.09} & 92.34 \\
 &  & wav2vec2LFT & 90.11 & 94.58 & 88.54 & 88.64 & 91.64 & 91.09 & 76.32 & 74.71 & 66.43 & 76.74 & 90.52 & \textbf{93.46} \\
 &  & hubertLFT & 89.91 & 94.57 & \textbf{90.92} & 85.92 & 91.69 & 87.55 & 80.93 & 76.15 & 65.35 & 74.84 & 90.78 & 93.43 \\
\midrule
LoRA & pTinyBERT & - & 86.91 & 90.23 & 74.97 & 53.94 & 87.82 & 75.71 & 77.12 & 75.47 & 50.00 & 52.10 & 75.31 & 84.09 \\
C-LoRA & pTinyBERT & - & 85.91 & 89.98 & 78.39 & 65.44 & 87.24 & 82.61 & 76.00 & 72.43 & 54.66 & 62.36 & 80.94 & 86.21 \\
\cmidrule[0.05pt](lr){1-3}
\multirow[t]{3}{*}{Transfer} & \multirow[t]{3}{*}{pTinyBERT} & whisperM & 84.07 & 88.75 & 86.82 & 86.52 & 86.95 & 87.00 & 71.37 & 79.08 & \textbf{83.61} & 83.07 & 87.11 & 87.77 \\
 &  & wav2vec2LFT & 83.04 & 88.42 & 86.18 & 84.52 & 86.21 & 86.69 & 73.06 & 76.84 & 80.53 & 79.95 & 85.88 & 86.73 \\
 &  & hubertLFT & 83.11 & 88.10 & 85.11 & 85.01 & 86.07 & 86.06 & 72.77 & 76.57 & 78.33 & 80.44 & 85.93 & 86.92 \\
\cmidrule[0.05pt](lr){1-3}
\multirow[t]{3}{*}{CoCo-LoRA} & \multirow[t]{3}{*}{pTinyBERT} & whisperM & 89.86 & 94.09 & 90.14 & 85.94 & 91.51 & 88.70 & 80.27 & 79.08 & 57.04 & 66.70 & 79.42 & 91.54 \\
 &  & wav2vec2LFT & 90.54 & 94.23 & 90.60 & 78.13 & 91.31 & 89.00 & 80.03 & 81.70 & 58.44 & 77.19 & 84.65 & 87.78 \\
 &  & hubertLFT & 90.51 & 93.78 & 89.11 & 75.13 & 91.59 & 86.15 & 80.09 & \textbf{81.87} & 60.90 & 75.91 & 90.02 & 91.07 \\
\midrule
LoRA & aMiniLM12 & - & 86.91 & 90.16 & 58.83 & 51.26 & 87.16 & 50.74 & 77.14 & 71.31 & 50.00 & 50.00 & 68.33 & 72.70 \\
C-LoRA & aMiniLM12 & - & 86.38 & 90.10 & 60.29 & 53.76 & 79.97 & 55.29 & 75.86 & 62.41 & 51.39 & 52.36 & 54.23 & 60.67 \\
\cmidrule[0.05pt](lr){1-3}
\multirow[t]{3}{*}{Transfer} & \multirow[t]{3}{*}{aMiniLM12} & whisperM & 82.68 & 83.32 & 80.24 & 81.81 & 84.06 & 81.81 & 72.29 & 78.50 & 78.86 & 80.40 & 82.44 & 80.56 \\
 &  & wav2vec2LFT & 81.28 & 84.12 & 80.56 & 78.22 & 83.01 & 80.08 & 69.64 & 75.65 & 72.78 & 75.07 & 80.36 & 82.11 \\
 &  & hubertLFT & 80.88 & 83.60 & 77.45 & 78.13 & 81.97 & 78.77 & 68.50 & 74.87 & 69.31 & 74.70 & 79.53 & 79.78 \\
\cmidrule[0.05pt](lr){1-3}
\multirow[t]{3}{*}{CoCo-LoRA} & \multirow[t]{3}{*}{aMiniLM12} & whisperM & \textbf{91.94} & \textbf{95.30} & 64.82 & 57.31 & 91.38 & 79.54 & \textbf{81.35} & 75.48 & 55.00 & 58.86 & 66.89 & 91.72 \\
 &  & wav2vec2LFT & 91.11 & 94.92 & 65.59 & 60.84 & 91.88 & 56.38 & 81.26 & 71.78 & 55.70 & 61.57 & 76.15 & 81.24 \\
 &  & hubertLFT & 91.59 & 95.13 & 55.46 & 56.76 & \textbf{92.17} & 75.05 & 79.96 & 70.35 & 55.14 & 63.50 & 72.13 & 89.06 \\
\midrule
LoRA & aDRoBERTa & - & 87.39 & 91.31 & 89.70 & 88.26 & 88.94 & 88.95 & 78.30 & 77.73 & 78.73 & 84.76 & 89.29 & 90.40 \\
C-LoRA & aDRoBERTa & - & 86.40 & 90.70 & 88.09 & 86.44 & 88.21 & 87.70 & 76.74 & 76.75 & 68.47 & 80.21 & 87.64 & 89.57 \\
\cmidrule[0.05pt](lr){1-3}
\multirow[t]{3}{*}{Transfer} & \multirow[t]{3}{*}{aDRoBERTa} & whisperM & 82.99 & 82.35 & 82.86 & 82.70 & 84.83 & 83.49 & 72.47 & 78.90 & 81.43 & 80.39 & 84.48 & 83.41 \\
 &  & wav2vec2LFT & 81.77 & 85.53 & 81.74 & 80.99 & 84.01 & 82.14 & 70.76 & 75.81 & 75.23 & 76.46 & 82.26 & 83.41 \\
 &  & hubertLFT & 81.73 & 84.80 & 80.38 & 79.58 & 83.01 & 81.52 & 68.97 & 75.77 & 68.52 & 75.51 & 81.40 & 82.23 \\
\cmidrule[0.05pt](lr){1-3}
\multirow[t]{3}{*}{CoCo-LoRA} & \multirow[t]{3}{*}{aDRoBERTa} & whisperM & 88.31 & 92.69 & 89.91 & 88.49 & 89.54 & 90.51 & 78.98 & 79.73 & 76.13 & 83.36 & 89.19 & 91.38 \\
 &  & wav2vec2LFT & 88.30 & 92.99 & 90.34 & \textbf{89.94} & 90.39 & 89.49 & 79.08 & 79.23 & 75.62 & 84.56 & 89.51 & 91.08 \\
 &  & hubertLFT & 88.93 & 92.89 & 88.74 & 82.09 & 90.08 & 90.25 & 78.43 & 81.11 & 77.65 & \textbf{85.93} & 89.39 & 90.46 \\
\bottomrule
\end{tabular}
}
\caption{AUC (\%) on a 5 fold cross-validation on segment-level predictions on parent data.}
\label{tbl_parent}
\end{table}

\begin{table}[!t]
\centering
\setlength\tabcolsep{2.75pt}
\scalebox{0.79}{
\hspace{-2.0cm}
\begin{tabular}{lllcccccccccccc}
\toprule
 % &  &  & \textbf{sentim} & \textbf{anger} & \textbf{fear} & \textbf{joy} & \textbf{sad} & \textbf{richness} & \textbf{coherence} & \textbf{rumination} & \textbf{worry} & \textbf{anxiousness} & \textbf{criticism} & \textbf{self- criticism} \\
 & & & \rot{\textbf{sentiment}} & \rot{\textbf{anger}} & \rot{\textbf{fear}} & \rot{\textbf{joy}} & \rot{\textbf{sadness}} & \rot{\textbf{richness}} & \rot{\textbf{coherence}} & \rot{\textbf{rumination}} & \rot{\textbf{worry}} & \rot{\textbf{anxiousness}} & \rot{\textbf{criticism}} & \rot{\textbf{self-criticism}} \\
Method & LLM & Audio Encoder &  &  &  &  &  &  &  &  &  &  &  &  \\
\midrule
LoRA & nRoBERTa & - & 93.88 & 87.36 & 73.03 & 83.36 & 81.42 & 83.99 & 78.82 & 51.05 & 64.98 & 66.81 & 93.12 & \textbf{93.82} \\
C-LoRA & nRoBERTa & - & 93.38 & 85.77 & 80.65 & 80.00 & 82.73 & 82.57 & 77.17 & 66.81 & 87.35 & 64.91 & 91.66 & 90.02 \\
\cmidrule[0.05pt](lr){1-3}
\multirow[t]{3}{*}{Transfer} & \multirow[t]{3}{*}{nRoBERTa} & whisperM & 93.19 & \textbf{88.37} & 87.90 & \textbf{86.63} & \textbf{84.88} & 84.22 & 82.34 & 83.46 & 87.93 & 80.90 & 92.52 & 92.58 \\
 &  & wav2vec2LFT & 93.03 & 87.48 & 73.82 & 83.59 & 82.31 & 83.54 & 80.51 & 73.44 & 86.72 & 75.24 & 91.31 & 91.70 \\
 &  & hubertLFT & 92.59 & 87.13 & 86.24 & 83.29 & 82.11 & 83.39 & 79.64 & 79.07 & 87.03 & 72.57 & 90.84 & 92.21 \\
\cmidrule[0.05pt](lr){1-3}
\multirow[t]{3}{*}{CoCo-LoRA} & \multirow[t]{3}{*}{nRoBERTa} & whisperM & 94.65 & 86.16 & \textbf{88.83} & 84.16 & 73.73 & 83.83 & 81.36 & 75.56 & 76.99 & 63.75 & 92.20 & 90.19 \\
 &  & wav2vec2LFT & 94.06 & 86.05 & 80.95 & 81.00 & 83.53 & 83.91 & 78.79 & 76.33 & 87.51 & 70.86 & 91.26 & 92.98 \\
 &  & hubertLFT & 94.10 & 86.95 & 83.55 & 81.32 & 79.61 & 83.25 & 78.36 & 84.15 & 87.50 & 65.72 & 93.27 & 91.05 \\
\midrule
LoRA & pTinyBERT & - & 90.92 & 53.15 & 50.25 & 51.92 & 50.30 & 72.80 & 72.50 & 60.17 & 50.03 & 52.86 & 50.00 & 50.00 \\
C-LoRA & pTinyBERT & - & 91.79 & 65.82 & 51.92 & 53.92 & 52.36 & 76.32 & 75.83 & 52.63 & 51.53 & 55.78 & 62.60 & 55.07 \\
\cmidrule[0.05pt](lr){1-3}
\multirow[t]{3}{*}{Transfer} & \multirow[t]{3}{*}{pTinyBERT} & whisperM & 90.86 & 86.08 & 84.29 & 85.66 & 83.69 & 85.09 & 82.22 & 83.73 & 83.40 & 79.47 & 90.32 & 89.98 \\
 &  & wav2vec2LFT & 90.78 & 84.83 & 74.03 & 81.31 & 79.98 & 82.67 & 79.56 & 75.44 & 82.01 & 71.68 & 87.18 & 88.43 \\
 &  & hubertLFT & 90.54 & 84.60 & 70.41 & 81.87 & 78.13 & 80.51 & 80.38 & 78.71 & 84.84 & 72.00 & 88.27 & 86.56 \\
\cmidrule[0.05pt](lr){1-3}
\multirow[t]{3}{*}{CoCo-LoRA} & \multirow[t]{3}{*}{pTinyBERT} & whisperM & \textbf{95.06} & 63.41 & 55.11 & 54.37 & 58.71 & 78.99 & 81.53 & 53.00 & 58.51 & 56.63 & 67.66 & 54.78 \\
 &  & wav2vec2LFT & 94.86 & 62.45 & 55.37 & 57.72 & 63.98 & 78.69 & 82.25 & 53.00 & 55.00 & 66.73 & 75.55 & 61.74 \\
 &  & hubertLFT & 94.64 & 79.64 & 53.00 & 56.72 & 64.00 & 79.37 & 81.32 & 54.50 & 53.54 & 60.57 & 75.86 & 56.03 \\
\midrule
LoRA & aMiniLM12 & - & 80.51 & 50.00 & 50.00 & 50.00 & 52.97 & 71.40 & 69.07 & 59.62 & 50.20 & 51.42 & 50.00 & 50.00 \\
C-LoRA & aMiniLM12 & - & 91.64 & 52.23 & 53.25 & 51.19 & 51.87 & 74.24 & 72.82 & 51.43 & 50.60 & 51.24 & 51.29 & 51.25 \\
\cmidrule[0.05pt](lr){1-3}
\multirow[t]{3}{*}{Transfer} & \multirow[t]{3}{*}{aMiniLM12} & whisperM & 75.96 & 79.29 & 80.12 & 80.81 & 75.73 & \textbf{85.11} & 81.98 & 68.45 & 75.09 & 79.84 & 79.47 & 83.45 \\
 &  & wav2vec2LFT & 81.76 & 75.10 & 72.52 & 71.75 & 64.75 & 78.89 & 79.10 & 62.52 & 66.45 & 72.02 & 77.42 & 77.76 \\
 &  & hubertLFT & 80.84 & 74.22 & 69.86 & 74.06 & 66.94 & 79.33 & 79.18 & 68.26 & 66.13 & 70.23 & 74.85 & 76.94 \\
\cmidrule[0.05pt](lr){1-3}
\multirow[t]{3}{*}{CoCo-LoRA} & \multirow[t]{3}{*}{aMiniLM12} & whisperM & 94.70 & 57.13 & 53.79 & 54.47 & 54.13 & 77.39 & 81.28 & 53.98 & 53.89 & 58.39 & 53.19 & 53.80 \\
 &  & wav2vec2LFT & 94.10 & 53.99 & 53.00 & 55.45 & 54.23 & 77.43 & \textbf{83.44} & 53.27 & 58.60 & 64.31 & 53.72 & 53.08 \\
 &  & hubertLFT & 94.63 & 55.76 & 53.37 & 55.27 & 56.06 & 77.52 & 83.06 & 53.69 & 53.00 & 59.03 & 55.67 & 53.43 \\
\midrule
LoRA & aDRoBERTa & - & 92.91 & 86.13 & 74.92 & 82.40 & 84.16 & 84.32 & 79.67 & 75.47 & 87.85 & 73.35 & 92.17 & 93.04 \\
C-LoRA & aDRoBERTa & - & 92.58 & 82.76 & 81.89 & 78.76 & 81.86 & 84.13 & 78.73 & 72.91 & 86.68 & 69.79 & 92.01 & 91.36 \\
\cmidrule[0.05pt](lr){1-3}
\multirow[t]{3}{*}{Transfer} & \multirow[t]{3}{*}{aDRoBERTa} & whisperM & 85.62 & 82.10 & 83.84 & 82.06 & 73.47 & 83.85 & 81.77 & 70.05 & 80.54 & \textbf{81.57} & 83.90 & 83.83 \\
 &  & wav2vec2LFT & 84.86 & 77.93 & 75.00 & 73.41 & 69.89 & 80.11 & 79.08 & 61.88 & 71.29 & 71.74 & 80.05 & 81.22 \\
 &  & hubertLFT & 84.47 & 78.35 & 70.11 & 76.10 & 71.19 & 79.01 & 78.18 & 68.18 & 65.96 & 71.80 & 78.71 & 76.86 \\
\cmidrule[0.05pt](lr){1-3}
\multirow[t]{3}{*}{CoCo-LoRA} & \multirow[t]{3}{*}{aDRoBERTa} & whisperM & 93.65 & 81.82 & 78.77 & 83.41 & 84.27 & 84.75 & 81.39 & \textbf{85.99} & 87.54 & 66.68 & \textbf{93.94} & 93.04 \\
 &  & wav2vec2LFT & 93.59 & 85.56 & 76.64 & 80.41 & 83.06 & 84.68 & 82.59 & 75.14 & \textbf{89.28} & 75.99 & 93.93 & 93.32 \\
 &  & hubertLFT & 93.64 & 84.14 & 81.71 & 79.34 & 83.32 & 84.92 & 81.49 & 74.64 & 88.53 & 70.37 & 93.58 & 91.78 \\
\bottomrule
\end{tabular}
}
\caption{AUC (\%) on a 5 fold cross-validation on segment-level predictions on offspring data.}
\label{tbl_offspring}
\end{table}

Across the offspring tasks as shown in Table \ref{tbl_offspring}, the main story is less `uniform wins everywhere' and more `where audio helps, it helps in a targeted way'. The clearest example is sentiment: CoCo-LoRA is consistently at or near the top for every text backbone, with especially large jumps when the underlying text model is weaker or unstable (e.g., paraTinyBERT and allMiniLM12 move from mediocre/near-chance sentiment to mid-high 90s AUC). This suggests the audio-conditioned posterior is doing something beyond generic regularization: it is providing a reliable contextual signal that stabilizes adaptation on a high-coverage label even when the text encoder alone struggles. By contrast, self-criticism is already extremely strong for nliRoBERTa with text-only LoRA, leaving limited headroom for any multimodal conditioning. In that regime, CoCo-LoRA behaves more like a competitive alternative than a consistent improver.

On the more acoustically sensitive but highly skewed affective labels (anger, fear, joy, sadness), the strongest results often come from the transfer baseline rather than CoCo-LoRA, especially for nliRoBERTa and paraTinyBERT with Whisper. This pattern is compatible with the idea that when the label is driven by salient prosodic cues, directly injecting audio into the classifier can over-specialize effectively, whereas CoCo-LoRA's KL-regularized variational pathway tends to be more conservative and may not exploit those rare-signal regimes as aggressively. The allDistilRoBERTa block is the closest to the best of both worlds behavior: CoCo-LoRA remains strong on sentiment and reaches the top result on rumination (with Whisper), while also matching or slightly surpassing transfer on criticism across encoders, indicating that the contextual uncertainty route can still extract useful audio information for selected downstream targets without relying on a large fusion head.

Overall, the results in Tables \ref{tbl_parent} and \ref{tbl_offspring} position CoCo-LoRA as a practical alternative to conventional multimodal fusion: it uses audio to guide adaptation and uncertainty rather than to inflate feature dimensionality, and this design yields broad gains with stable behavior across model choices.

\section{Conclusion}

CoCo-LoRA provides a lightweight yet expressive route to multimodal, uncertainty-aware adaptation for speech-centered prediction tasks. Instead of treating audio as another high-dimensional feature stream to be fused into a classifier, we use it as a reliability/context signal that shapes a contextual variational posterior in the low-rank adapter space. By projecting a pooled audio embedding once and refining it through small layer-wise heads, CoCo-LoRA achieves global-to-local, depth-specific modulation while keeping the backbone frozen and confining stochasticity to a compact rank-space latent. This preserves the scalability advantages of PEFT and yields audio-sensitive, heteroscedastic uncertainty that better reflects external acoustic variability (e.g., noise, channel effects, speaking style) than unimodal uncertainty-aware adapters.

Across a broad set of tasks, language-model backbones, and audio encoders, our results show that CoCo-LoRA consistently matches or improves upon strong text-only PEFT baselines and frequently surpasses a conventional multimodal transfer baseline based on feature fusion. The most reliable improvements appear on high-coverage labels, where conditioning the adapter distribution on audio context provides a stable and effective signal for adaptation. In contrast, on some highly imbalanced, rare-positive labels, fusion baselines can occasionally dominate, suggesting that direct head-level audio injection may over-specialize to salient acoustic correlates, while CoCo-LoRA’s KL-regularized variational objective favors more conservative updates when evidence is sparse. Overall, these patterns support our central claim: using audio to modulate uncertainty and adaptation, rather than to expand representations via fusion, offers a robust and parameter-efficient alternative for low-resource multimodal prediction.

As future work, the current global-to-local pathway could be made adaptive (e.g. by learning which layers should rely more on audio context, or by using mixture-of-experts/gating over layer heads), to better match task-specific acoustic reliance without increasing parameter cost substantially. Additionally, future studies can extend the framework to richer audio conditioning (e.g., multiple pooled summaries or time-aware context with cross-modal attention) while retaining the core constraint that multimodality should inform reliability and adaptation rather than require heavy multimodal fusion.

\bibliographystyle{splncs04}
\bibliography{references}

\end{document}